\documentclass[review]{elsarticle}

\usepackage{lineno,hyperref}

\journal{Artificial Intelligence}

\usepackage[utf8]{inputenc}
\usepackage{tikz}
\usepackage{amsfonts}
\usepackage{amsmath}
\usepackage{mathtools}
\usepackage{colortbl}
\usepackage{diagbox}
\usepackage{multirow}
\usepackage{fixltx2e}
\usepackage{graphicx}

\usepackage{setspace}
\usepackage[T1]{fontenc}
\usepackage{lmodern}
\usepackage[babel=true]{microtype}

\begin{document}
	
	\begin{frontmatter}
		
		\title{CTRAN: CNN-Transformer-based Network for Natural Language Understanding}

		\author[mymainaddress]{Mehrdad Rafiepour\corref{mycorrespondingauthor}}
		\ead{Mehrdad.Rafie.p@gmail.com}
		\author[mymainaddress]{Javad Salimi Sartakhti\corref{mycorrespondingauthor}}
		\ead{salimi@kashanu.ac.ir}
		\address[mymainaddress]{Computer and Electrical Engineering Department, University of Kashan, Kashan, Iran.}
		\cortext[mycorrespondingauthor]{Corresponding author}
		\begin{abstract}
			Intent-detection and slot-filling are the two main tasks in natural
			language understanding. In this study, we propose CTRAN, a novel encoder-decoder CNN-Transformer-based architecture for intent-detection and slot-filling. In the encoder, we use BERT, followed by several convolutional layers, and rearrange the output using window feature sequence. We use stacked Transformer encoders after the window feature sequence. For the intent-detection decoder, we utilize self-attention followed by a linear layer. In the slot-filling decoder, we introduce the aligned Transformer decoder, which utilizes a zero diagonal mask, aligning output tags with input tokens. We apply our network on ATIS and SNIPS, and surpass the current state-of-the-art in slot-filling on both datasets. Furthermore, we incorporate the language model as word embeddings, and show that this strategy yields a better result when compared to the language model as an encoder. 
		\end{abstract}
		
		
	\end{frontmatter}

	\section{Introduction}
		With the rapid growth of smartphones, digital assistants are becoming more involved in our life. They make routine tasks easier and improve our quality of life. Dialogue systems are the backbone of digital assistants. Moreover, goal-driven dialogue systems are used in banking, website customer support, and travel guidance. Natural language understanding (NLU), a critical part of dialogue systems, is a prime issue in human-machine interaction. NLU has two sub-tasks. The first is to understand the intent that the user has in mind, and the second is to extract the semantic information from the sentence. The first task is called intent-detection (ID), and the latter is referred to as slot-filling (SF). ID can be defined as a classification problem, and SF can be considered as a sequence labeling problem. Table \ref{atis-dataset} shows an example of the ATIS dataset, which contains the user's question, intent, and target labels.\\
		We spotted three significant issues affecting the performance of the existing models.
		Recurrent neural networks (RNNs) like long short-term memory (LSTM)\citep{lstm:hochreiter1997} have been widely adopted in the structure of networks proposed to solve SF and ID tasks\citep{e:2019,kane:2020}. The First problem is that RNNs cannot capture the deep bidirectional meaning of a sentence since their output vector is a concatenation of two unidirectional RNNs\citep{BERT:devlin2019}. Transformer \citep{transformer:vaswani2017} addressed this problem with self-attention and positional embeddings.
		The second issue is that previous works used convolutional layers with kernel sizes of 2 or more \citep{Wang:18,zhou:15clstm}. We argue that the output of the convolutional layer may still have the embedding of the word present, but the one-to-one relation of input to output tokens needs to be explicit. The third issue is that BERT is mostly used as an encoder, and the embeddings are not further encoded in most existing NLU models. We investigate the best strategy for using a language model in the architecture of an NLU model. In this paper, we propose CTRAN, a novel CNN-Transformer-based encoder-decoder network. For the encoder, we experiment with both BERT and ELMo as our word embedding. We use CNN on word embeddings and restructure its output using window feature sequence (WFS). The final part of the encoder is stacked Transformer encoders. The decoder uses self-attention and a linear layer to classify the user's intent. For the SF task, we propose the aligned Transformer decoder followed by a fully connected layer.
		In short, our contribution has four folds.
		\begin{table*}
			\centering
\begin{tabular}{|c|ccccc|}
\hline
\textbf{user's question} & \multicolumn{1}{c}{what} & \multicolumn{1}{c}{is} & \multicolumn{1}{c}{the} & \multicolumn{1}{c}{abbrevation} & d10              \\ \hline
\textbf{target label}    & \multicolumn{1}{c}{O}    & \multicolumn{1}{c}{O}  & \multicolumn{1}{c}{O}   & \multicolumn{1}{c}{O}           & B-aircraft\_code \\ \hline
\textbf{target intent}   & \multicolumn{5}{c|}{atis\_abbreviation}  \\
\hline
\end{tabular}
			\caption{\label{atis-dataset}
				Example input, target label and intent in ATIS dataset. 
			}
		\end{table*}
		\begin{enumerate}
			\item We propose a novel joint intent-detection and slot-filling architecture for natural language understanding, achieving state-of-the-art for slot-filling on both ATIS and SNIPS. 
			\item We introduce alignment in the Transformer decoder to keep the one-to-one relationship between input tokens to output tags.
			\item We propose using convolutional layers with the kernel size of 1 to preserve the original one-to-one relationship while fusing the word embedding with adjacent words.
			\item We use the language model as word embeddings and show that this strategy performs better when compared to employing the language model as an encoder.
		\end{enumerate}
		The rest of this paper is ordered as follows: We discuss related works in section 2. Section 3 describes our proposed network. In section 4, we specify our experiment setup. Section 5 contains our results and analysis, and in Section 6, we conclude our study and discuss future work.
	\section{Related Work}
		Although there are several statistical models, neural networks have proven to be more accurate. In the past, ID and SF tasks were carried out separately \citep{mesnil:2015,yang-etal-2016-hierarchical}. However, recent studies have shown that the joint training of two tasks yields a better result\citep{Wang:18,goo-etal-2018-slot,zhang-2018-joint,Qin:2019}.\\
		Traditional models mostly used pre-trained fixed vector embeddings such as GloVe \citep{glove} and Word2Vec \citep{mikolov2013efficient} for the word representation. 
		CoBiC uses CNN to process the embeddings of the input tokens \citep{kane:2020}. The output of the CNN is fed to a bi-directional LSTM, and then a conditional random field (CRF) layer is used to generate the output labels. CoBiC uses the hidden state of the last unit to produce intents.
		\citet{yang:2021} introduced AISE, using a bi-directional LSTM as a shared encoder. For ID, AISE uses multi-head attention pooling. Additionally, \citet{yang:2021} introduced the position-aware multi-head masked attention mechanism for their SF decoder.
		\citet{e:2019} proposed Reinforce Vector as a solution for intent-slot integration. They utilize a shared Bi-LSTM as the encoder. They concatenate the last hidden state of the encoder with their novel Intent Reinforce Vector followed by softmax for intent prediction. Furthermore, Slot Reinforce Vector is concatenated with the encoder's hidden state, which is then fed to a CRF layer for slot generation. 
		\\
		In recent years, pre-trained language models have become progressively adopted as they are proven to be beneficial for many downstream tasks \citep{Qiu2020}. NLU models either use the language model as word embeddings or use the structure of the language model as their encoder and only propose a decoder. As of latter, the model relies primarily on fine-tuning the language model. For example, \citet{chen:2019} used BERT as an encoder, harnessed [CLS] token with a softmax layer for ID and rest of the token outputs with a softmax for generation of target slots. \citet{Wang:2020} introduced SASGBC, which utilized BERT to embed words into vectors. BERT's [CLS] token is then integrated into each slot, and then applied the self-attention mechanism after it. In the end, they used a CRF layer to produce final labels for each slot.  
		In the case of language model as word embeddings, \citet{huang:2020} introduces multi-view encoder to be used after BERT, which is consisted of several encoders, namely position-wise encoder, local encoder, global encoder, and time-series encoder. Furthermore, they propose federated learning, meaning the encoder share parameters between 4 different tasks (ID, SF for Snips and ATIS).
		\citet{Qin:2021} restructured encoder-decoder into encoder - Co-Interactive module - decoder to build bidirectional connection between SF and ID. They also used BERT as a word embedding, followed by coding.
	\section{CTRAN}
\begin{figure*}
			\includegraphics{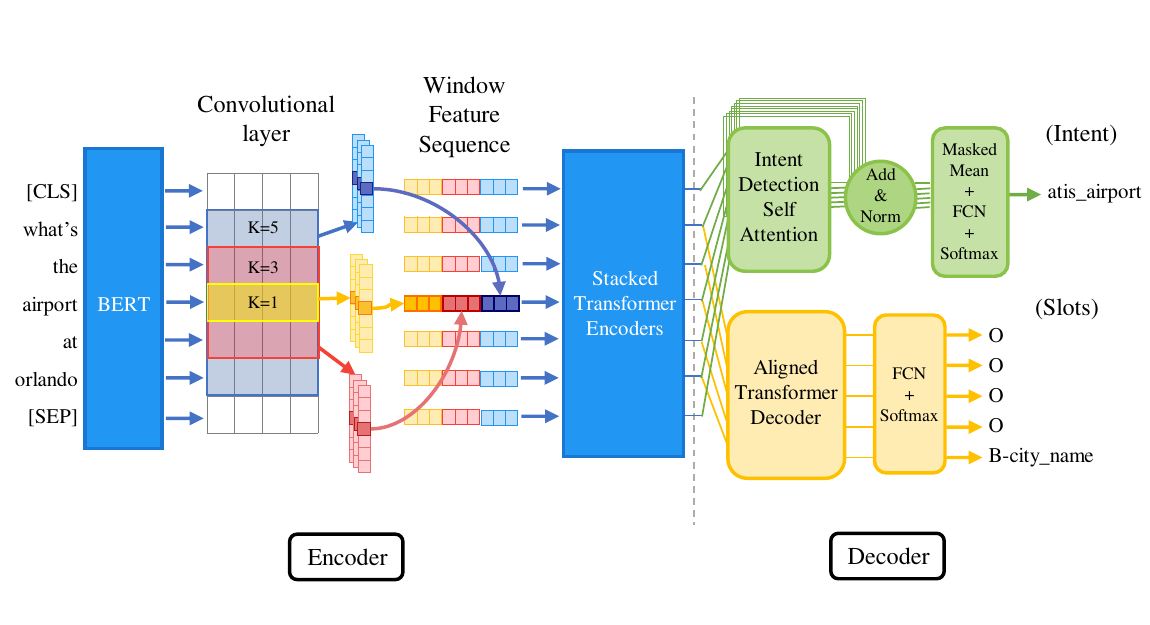}
			\caption{\label{model_architecture}CTRAN's architecture. Note that it is required for BERT to add the [CLS] and [SEP] tokens to the beginning and end of the sentence.}
		\end{figure*}

		In this section, we explain different components of our proposed joint network. Figure \ref{model_architecture}  shows a detailed overview of our proposed encoder-decoder architecture. CTRAN comprises three main components: A shared encoder, an SF decoder, and an ID decoder. SF and ID tasks use the same encoder, meaning both tasks use the vector produced by the encoder, and the subsequent losses are summed before calculating the gradients. Hence, CTRAN benefits from joint learning.
		\subsection{Shared Encoder}
			The proposed encoder consists of a pre-trained language model, CNN, window feature sequence structure, and stacked Transformer encoders.
			First, as shown in Figure \ref{model_architecture}, tokens are fed to BERT, which outputs a contextualized embedding that we further process in our network. Then, convolution operation with different kernel sizes takes place over the embedded words. It is usual to use operations like max-pooling after convolutions. However, intending to conserve the original order of tokens, we used WFS, in which we transpose and concatenate the values related to each word into a single vector. The final piece is stacked Transformer encoders, which we use to capture the new representation provided by the WFS structure. Following subsections describe the different parts of our shared encoder.
		
	\subsubsection{Pre-Trained Language Model}
		Pre-trained language models such as ELMo \citep{elmo:peters2018} and BERT have proven to be effective in improving the performance of the downstream NLP tasks \cite{ethayarajh-2019-contextual}. We employ both language models in our architecture and compare the results. BERT consists of several bi-directional Transformer encoders. 
		$BERT_{base}$ has 12 and $BERT_{large}$ is made of 24 layers. 
		Input to the BERT is the sum of WordPiece token embeddings \cite{wordpiece}, positional embedding, and segment embedding. The latter does not affect our case since we only provide one sentence to the model.
		For the given input sentence $X=\left [x_{1},x_{2},x_{3},...,x_{L}\right ]$, BERT provides contextualized embeddings $H=\left [h_{1},h_{2},h_{3},...,h_{L}\right ]$.
		Note that WordPiece tokenization treats punctuations as separate tokens, breaking words containing punctuation marks into several tokens. This issue causes token clutter and out-of-order tokens. We solved this issue by removing all punctuation marks. ELMo is comprised of CNN for character embedding and bi-directional LSTM, which are trained on a large corpus. The main idea behind ELMo is that word embeddings are a function of other words in the sentence, meaning the embedding of a word depends on the context of the sentence in which it is used.
		ELMo is a sequence-to-sequence model that uses non-contextual character-based word representation. The benefit of such representation is the ability to deal with word misspellings and unseen words.
		\subsubsection{Convolutional Layer}
		In \citep{Wang:18} and \citep{zhou:15clstm} CNN was used to cover the shortcomings of LSTM. We use Transformer encoder instead of an LSTM network. While Transformer rectifies said flaws, we argue that CNN on Transformer can still be beneficial. We use CNN to put an emphasis on the meaning of adjacent tokens. We utilize several convolutional layers with different kernel sizes and initial values to fuse the embedding of each token with neighboring tokens. 
		Assume that $d$ represents embedding dimension. If input sentence has $L$ tokens, the input sentence would be defined as $x\in\mathbb{R}^{L\times d}$ and embedding vector for the $i$-th token is shown by $x_{i}\in \mathbb{R}^{d}$. Let $k$ be a filter size and $f\in\mathbb{R}^{k\times d}$ denote a filter-map. Then for each position $i$ in the sentence, there is a window $w_{i}$ containing k token vectors as  $w_{i}=\left[x_{i},x_{i+1},x_{i+2},...,x_{i+k-1}\right]$.
		$\odot$ is element-wise multiplication, $b$ is bias and $A$ is an activation function. Hence, the convolution operation $c_{i}$ for each window $w_{i}$ is given by:
		\begin{equation}\label{cnnoperation} 
			c_{i}=A\left(w_{i}\odot f+b\right)
		\end{equation}
		This operation is done over all available indexes $i$, so the result of $f$ convolving over $x$ is represented by $C=\left[c_{1},c_{2},c_{3},...,c_{i},...,c_{L}\right]$.
		Note that we use zero padding, so the convolutional layer's input and output always have the same length.
	
	\subsubsection{Window Feature Sequence}
	It is usual to use pooling operations after a convolutional layer, but the discontinuous sampling will lose some of the information, and subsequently, such operations will destroy sequential data. We use WFS to retain the original order of words.\\
	Let $V$ be the total number of filters with different kernel sizes and initial values for the convolutional layer, and $j$ denotes the index referring to $j$-th filter. $R_{i}$ defined in Equation \ref{cnnfinal}, shows final embedding vector of the word $i$ and $C_{i}^{j}$ indicates $i$-th element of the $j$-th filter,
	\begin{equation}
		\label{cnnfinal}
		R_{i}=C_{i}^{1} \oplus C_{i}^{2} \oplus C_{i}^{3} \oplus ... \oplus C_{i}^{j} \oplus ... \oplus C_{i}^{V}
	\end{equation}
	Where $\oplus$ represents concatenation, therefore $R_{i}$ concatenates all of the convolutional values belonging to index $i$ for all of the filter-maps.
	This operation is done over all indexes until all the word vectors are produced for the input words. These vectors are then fed to stacked Transformer encoders to produce the final word representations.
	\subsubsection{Transformer Encoder} 
	Directly accessing all of the positions in the sentence helps the Transformer encoder overcome information loss.
	Stacked Transformer encoders consist of several layers where each layer's output is used as the next layer's input. We utilize stacked Transformer encoders to generate a new representation for the words that incorporates an emphasis on the neighboring words. We use 2 layers of Transformer encoders which is smaller relative to BERT, in the light that this layer of the network only has one objective and general meaning of the word is provided through BERT. Transformer encoder is the final part of our shared encoder architecture. The vector provided by this layer will be used in both ID and SF decoders.
	\subsection{Intent-Detection Decoder}
		As shown in Figure \ref{model_architecture}, ID decoder consists of a self-attention layer with residual connection and layer normalization, followed by a linear feed-forward network with softmax applied for classification. Accordingly, output for the ID self-attention layer is determined by following:
		\begin{equation}
			\label{decoder_residual}
			S = e + LN(  MultiHead(e)  )
		\end{equation}
		Where $e\in \mathbb{R}^{L\times d_{model}}$ is encoder output, and $LN$ denotes Layer-Normalization. Moreover, $MultiHead(e)$ denotes the $MultiHead$ attention function defined by \citet{transformer:vaswani2017} where $Q, K, V$ are transformed from $e$. Furthermore, S represents output of the operation.
		Finally, probability distribution between intents is computed with Equation \ref{decoder_id}.
		\begin{equation}
			\label{decoder_id}
			P_{intent} = softmax( W\cdot S + b )
		\end{equation}
		Note that $b$ and $W$ are parameters to be learned at training time.
	\subsection{Slot-Filling Decoder}
		The proposed SF decoder is made of two parts: an aligned Transformer decoder and a linear layer to bring down the dimension to the number of tags. Since the input to output tags have a one-to-one relationship, The Transformer decoder needs to be aligned. The proposed alignment happens in the cross-attention segment. In this section, keys and values are generated from the encoder's output which we call memory. A matrix with the size of $(T, S)$ is needed for masking the memory, where $T$ and $S$ are the target and source length. Since input and output have the same shape, the mask will be in the shape of $(S, S)$. We aligned the Transformer decoder by applying a zero diagonal mask for the memory. Hence, every position in the memory except for $t = s$ is masked out where s is the row index, and t is the column index. Thus, all the keys related to positions other than the position of generating token would change to zero. Consequently, the Transformer decoder only considers the corresponding embedding vector for each predicting token. In order for masking to be applied on $MultiHead$, it is altered to equation \ref{Attention}. In this equation, $Q\in\mathbb{R}^{S\times d_{k}}, K\in\mathbb{R}^{S\times d_{k}}, V\in\mathbb{R}^{S\times d_{v}}$ and $M\in\mathbb{R}^{S\times S}$ represent query, key, value and mask matrices, and $d$ denotes dimension.
		\begin{equation}\label{Attention}
			MultiHead_{Masked}(Q, K, V, M) = softmax(\frac{Q . K^{T}}{\sqrt{d_{k}}}+M) . V
		\end{equation}
		Similar to the original Transformer decoder, the aligned Transformer decoder can be divided into three segments based on functionality.
		Considering Query, Key, and Value transformed from target token embeddings $D\in\mathbb{R}^{S\times d_{model}}$, first segment which utilizes self-attention to perceive the relations between previously generated tokens is defined as following:
		\begin{equation}\label{transformer_decoder}
			C^{d}=D+LayerNorm(MultiHead_{Masked}(Q,K,V,M_{upper}))
		\end{equation}
		Note that $M_{upper}$ is a strictly upper diagonal matrix where all entries above the main diagonal are $-inf$. This mask changes the attention score for positions ahead to zero. Thus, the attention mechanism will not be dependent on future tokens. \\ 
		Considering $K^{e}\in\mathbb{R}^{S\times d_{k}}$, $V^{e}\in\mathbb{R}^{S\times d_{v}}$ as key and value transformed from memory, the second segment utilizes cross-attention to use information from the encoder is given by:
		\begin{equation}\label{transformer_decoder2}
			F^{d}=C^{d}+LayerNorm(MultiHead_{Masked}(C^{d},K^{e},V^{e},M_{zerodiag}))
		\end{equation}
		Note that $M_{zerodiag}$ is a zero diagonal matrix in which the main diagonal is zero and all other entries are $-inf$. When the softmax function is applied, the off-diagonal entries would change to zero. Consequently, only the value vector for corresponding positions in the memory is added to the $C^{d}$ matrix.\\
		With $FFN$ representing the position-wise feed-forward network, the output of the aligned Transformer decoder is $O^{d}=LayerNorm(FFN(F^{d}))+F^{d}$.
		Finally, the tag probability distribution is calculated by $P=softmax(W\cdot O^{d}+b)$ 
		where both W and b are trainable parameters.
		
	\section{Experiments and Setup}
		In this section, the datasets used to evaluate CTRAN are introduced, and then, the hyper-parameters in our experiments are specified. 
		\subsection{Datasets}
			We conduct experiments on two well-known datasets, namely ATIS and SNIPS, to measure the performance of our proposed network. We train the network on the train set, tune it on the dev set and report the final results on the test set. Some of the tags in the dev and test sets may not appear during training phase. In that case, we have assigned them the Unknown class.\\
			ATIS benchmark \citep{atis:hemphill1990} which is widely adopted in the evaluation of NLU models, is a dataset regarding people asking questions about flight information and reservations. It contains 4478, 500, and 893 for train, dev, and test set, respectively. Moreover, ATIS has 127 slot types and 21 intents.\\
			SNIPS benchmark \citep{snips:coucke2018} is a dataset from SNIPS personal assistant containing 13084, 700, and 700 utterances for training, test, and dev sets, respectively. Furthermore, it consists of 7 types of intents and 72 types of slots.
			In contrast to ATIS, SNIPS is more complicated, containing several domains and a more extensive vocabulary.
			\begin{table}
				
\centering
\arrayrulecolor{black}
\begin{tabular}{|l|l|l|l|} 
\hline
Layer\textbackslash{}Parameter                 & $\alpha$                                                                   & $\gamma$ & $\rho$  \\ 
\hline
BERT~~$\begin{matrix}base\\large \end{matrix}$ & \begin{tabular}[c]{@{}l@{}}$10^{\text{-}4}$\\$10^{\text{-}5}$\end{tabular} & $0.96$   & $0.1$   \\ 
\arrayrulecolor[rgb]{0.753,0.753,0.753}\hline
ELMo                                           & $10^{\text{-}4}$                                                           & $0.96$   & $0.5$   \\ 
\hline
CNN \& WFS                                     & $10^{\text{-}3}$                                                           & $0.96$   & $0.0$     \\ 
\hline
Transformer Encoder                            & $10^{\text{-}4}$                                                           & $0.96$   & $0.1$   \\ 
\hline
ID Decoder                                     & $10^{\text{-}4}$                                                           & $0.96$   & $0.5$   \\ 
\hline
SF Decoder                                     & $10^{\text{-}4}$                                                           & $0.96$   & $0.5$   \\
\arrayrulecolor{black}\hline
\end{tabular}

				\caption{\label{table_rates} Hyper-parameters used in training phase. Learning-rate $\alpha$, scheduler rate $\gamma$ and dropout probability $\rho$ used for each layer while training our model.}
			\end{table}
		\subsection{Experimental Settings}
			Hyper-parameters have an essential role in the performance of a neural network model. Since each component of CTRAN uses a distinct kind of neural network, we used different learning rates $\alpha$ with AdamW optimizer \citep{adamw:Loshchilov} for each layer. We adjust the learning rate with the StepLR scheduler, which reduces the learning rate of each parameter by decrease rate $\gamma$. Also, dropout \citep{dropout:Srivastava} is applied to the layers to decrease the amount of overfitting. We also found that using gradient clipping \citep{clipping:mikolov} with the value of 0.5 helps the final performance of the model. Table \ref{table_rates} shows the $\alpha$, $\gamma$ and $\rho$ for each layer. Also, the batch size in all datasets was 16.
			We tried 1, 2, 3, [1,3], [1,3,5], [2,3,5], and [1,2,3,5] as kernel sizes where brackets denote multiple kernel sizes used. The cumulative filter count is always 512, and the number of filters is evenly spread between different kernels.
			We ran our model 10 times, each having 50 epochs, and recorded the best results for each task. We reported the median value as the result for each experiment.

	\subsection{Results and Analysis}
		\begin{table}

\centering
\arrayrulecolor[rgb]{0.753,0.753,0.753}
\begin{tabular}{!{\color{black}\vrule}l!{\color{black}\vrule}c|c!{\color{black}\vrule}} 
\arrayrulecolor{black}\hline
Model                                                                                          & ID acc &  SF f1  \\ 
\hline
\begin{tabular}[c]{@{}l@{}}SASBGC$\beta$\\ \citep{Wang:2020}\end{tabular}                             &  98.21 &  96.69  \\ 
\arrayrulecolor[rgb]{0.753,0.753,0.753}\hline
\begin{tabular}[c]{@{}l@{}}Joint Bert$\beta$\\ \citep{chen:2019} \end{tabular}                        &  97.50 &  96.10  \\ 
\hline
\begin{tabular}[c]{@{}l@{}}CNN-BLSTM-Aligned\\ \citep{Wang:18}\end{tabular}                    &  97.17 &  97.76  \\ 
\hline
\begin{tabular}[c]{@{}l@{}}CharEmbed+CNN-LSTM-CRF\\ \citep{Firdaus:2019} \end{tabular}         &  99.09 &  97.32  \\ 
\hline
\begin{tabular}[c]{@{}l@{}}Elmo+BiLSTM+CRF\\ \citep{Siddhant:2019} \end{tabular}               &  97.42 &  95.62  \\ 
\hline
\begin{tabular}[c]{@{}l@{}}Bi-directional Interrelated\\ \citep{e:2019}\end{tabular}           &  97.76 &  95.75  \\ 
\hline
\begin{tabular}[c]{@{}l@{}}Co-interactive Transformer$\beta$ \\ \citep{Qin:2021} \\\end{tabular} &  98.00 &  96.10  \\ 
\hline
\begin{tabular}[c]{@{}l@{}}Federated Learning$\beta$\\ \citep{huang:2020} \end{tabular}          &  98.28 &  96.41  \\ 
\hline
\begin{tabular}[c]{@{}l@{}}CoBiC\\ \citep{kane:2020} \end{tabular}                             &  99.43 &  97.82  \\ 
\arrayrulecolor{black}\hline
\textbf{CTRAN$\beta$}                                                               &\bf{98.07}&\bf{98.46}\\
\hline
\end{tabular}

			\caption{\label{table_final_atis} A comparison between other well-known models and the proposed model on ATIS dataset. $\beta$ denotes BERT used.}
		\end{table}	
		As the models reach near 100\% accuracy, improvements become smaller and harder to achieve. Table \ref{table_final_atis} compares our results with current well-known best-performing models on ATIS. CTRAN+BERT\textsubscript{large} outperforms the current state-of-the-art model on ATIS with 0.64\% improvement and sets a new record in the SF task. Furthermore, for the intent-detection task, our proposed structure showed improvement over \citet{Wang:18} which used CNN-WFS similar to us. 
		\begin{table}

\centering
\arrayrulecolor[rgb]{0.753,0.753,0.753}
\begin{tabular}{!{\color{black}\vrule}l!{\color{black}\vrule}c|c!{\color{black}\vrule}} 
\arrayrulecolor{black}\hline
Model                                                                                            & ID acc & SF f1  \\ 
\hline
\begin{tabular}[c]{@{}l@{}}SASBGC$\beta$\\ \citep{Wang:2020}\end{tabular}                        &  98.86 & 96.43  \\ 
\arrayrulecolor[rgb]{0.753,0.753,0.753}\hline
\begin{tabular}[c]{@{}l@{}}Joint Bert$\beta$\\ \citep{chen:2019} \end{tabular}                   &  98.60 & 97.00  \\ 
\hline
\begin{tabular}[c]{@{}l@{}}CM-net$\beta$\\ \citep{liu:2019} \end{tabular}                        &  99.32 & 97.31  \\ 
\hline
\begin{tabular}[c]{@{}l@{}}Masked Graph+CRF$\beta$\\ \citep{Tang:2020} \end{tabular}             &  99.70 & 97.20  \\ 
\hline
\begin{tabular}[c]{@{}l@{}}Elmo+BiLSTM+CRF\\ \citep{Siddhant:2019} \end{tabular}               	 &  99.29 & 93.90  \\ 
\hline
\begin{tabular}[c]{@{}l@{}}Co-interactive Transformer$\beta$ \\ \citep{Qin:2021} \\\end{tabular} &  98.80 & 97.10  \\ 
\hline
\begin{tabular}[c]{@{}l@{}}Federated Learning$\beta$\\ \citep{huang:2020} \end{tabular}          &  99.33 & 97.20  \\ 
\hline
\begin{tabular}[c]{@{}l@{}}AISE$\beta$\\ \citep{yang:2021} \end{tabular}        				 &  98.70 & 97.20  \\ 
\arrayrulecolor{black}\hline
\textbf{CTRAN$\beta$}                                                              &\bf{99.42}&\bf{98.30}\\
\hline
\end{tabular}

			\caption{\label{table_final_snips} Comparing the proposed model with other well-known models on SNIPS dataset.$\beta$ denotes BERT used.}
		\end{table}
		To inspect the generalization capacity of our model, we also applied CTRAN on SNIPS. Table \ref{table_final_snips} compares well-known models with CTRAN. We improve near 1\% upon previous state-of-the-art for the SF task on SNIPS. Also, our intent-detection accuracy is near current state-of-the-art but not surpassing it. In the following subsections, we will investigate how much each idea improved CTRAN.
	\subsubsection{The Effect of Convolutional Layer With Transformer Encoder}
		Since previous models used CNN to treat the inherent deficiency of the LSTM, the adoption of Transformer encoder may alleviate the importance of the convolutional layer. Our goal for incorporating CNN-WFS with Transformer encoder is to fuse the adjacent token embeddings. Table \ref{table_cnn_wfs_te} compares the performance of a Transformer encoder with our proposed CNN-WFS-Transformer Encoder. For both ATIS and SNIPS the SF f1 and ID accuracy improved with CNN-WFS-Transformer encoder architecture. The results corroborate our intuition about the importance of adjacent tokens.
		\begin{table}
			\def\arraystretch{1.2}

\centering
\begin{tabular}{|c|c|c|c|c|} 
\hline
\multicolumn{1}{|r|}{Dataset} & \multicolumn{2}{c|}{ATIS} & \multicolumn{2}{c|}{SNIPS}  \\ 
\cline{2-5}
\multicolumn{1}{|l|}{Model}   & ID acc & SF f1            & ID acc & SF f1              \\ 
\hline
CNN-WFS-TE                    & 97.95  & 98.39            & 99.42  & 98.21              \\ 
\hline
TE only                       & 97.88  & 98.36            & 99.01  & 97.91              \\
\hline
\end{tabular}

			\caption{\label{table_cnn_wfs_te} A comparison between CNN-WFS-TE and Transformer encoder only. Experiments were done on $BERT_{base}$.}
		\end{table}
	\subsubsection{Kernel Size}
		Kernel size specifies the width in which local semantic information is extracted. Furthermore, kernel size in convolutions can be viewed as n-grams, in that a size of 2 is similar to bigram. To the best of our knowledge, the kernel size of 1 was not used in similar papers which adopted convolutional layers for NLU. We include 1, to explicitly keep the token embeddings of each word while bringing down the embedding dimension to match the other kernel sizes. This helps the model fuse adjacent word embeddings while keeping explicit embeddings of each word. Table \ref{table_cnn_kernels} shows the effect of different kernel sizes on the performance of CTRAN. Amongst single kernel sizes, 1 performs the best for the SF task. The reason is that a kernel size of 1 saves the unigram embeddings, which does not interfere with the one-to-one relation of the input to output tags. Moreover, comparing [2,3,5] to [1,2,3,5] confirms our previous statement. Based on our results, we can conclude that using multiple kernel sizes is better than a single kernel size. Furthermore, we can conclude that using multiple kernel sizes is better than a single kernel size. At the endnote, [1,2,3,5] achieves the best accuracy in SF and ID tasks.
		\begin{table}
			\def\arraystretch{1.4}
\centering
\begin{tabular}{|l|c|c|c|c|} 
\hline
\multicolumn{1}{|c|}{~Dataset} & \multicolumn{2}{c|}{ATIS} & \multicolumn{2}{c|}{SNIPS}  \\ 
\cline{2-5}
Size                           & ID acc & SF f1            & ID acc & SF f1              \\ 
\hline
1                              & 97.61  & 98.39            & 98.69  & 98.14              \\ 
\hline
2                              & 97.73  & 98.37            & 98.84  & 98.13              \\ 
\hline
3                              & 97.84  & 98.35            & 98.84  & 98.13              \\ 
\hline
{[}1,3]                        & 97.84  & 98.40            & 98.98  & 98.15              \\ 
\hline
{[}1,3,5]                      & 97.95  & 98.43            & 99.13  & 98.25              \\ 
\hline
{[}1,2,3,5]                    & 98.07  & 98.46            & 99.13  & 98.30              \\
\hline
{[}2,3,5]                      & 98.07  & 98.38            & 98.98  & 98.20              \\
\hline
\end{tabular}

			\caption{\label{table_cnn_kernels}The result of different kernel sizes in the convolutional layer. Bracket denotes multiple kernel sizes.}
		\end{table}
	\subsubsection{The Transformer Decoder Alignment}
		A regular Transformer decoder does not use any key-maskings other than the padding masks in cross-attention segment; meaning, in the generation of each target token, the key vector transformed from all context positions are used in the computation of the cross-attention. In contrast, in an aligned Transformer decoder, only the key generated from the context corresponding to each predicting token affects the cross-attention output. In order to corroborate the effectiveness of the proposed alignment, we also combined CTRAN's encoder with a regular Transformer decoder, and in another experiment, with an aligned LSTM with attention mechanism as described by \citet{Wang:18}. Table \ref{table_notaligned} compares the SF f1 of the three experiments in which the proposed alignment shows an average of 0.9\% improvement over the regular Transformer decoder. Furthermore, comparing the proposed aligned Transformer decoder with the \citet{Wang:18} SF decoder shows 0.1\% improvement, stating it is overall a better architecture for the SF task.
		\begin{table}
			\def\arraystretch{1.4}
\centering
\begin{tabular}{|l|c|c|} 
\hline
\multicolumn{1}{|r|}{Dataset}   & \multirow{2}{*}{ATIS} & \multirow{2}{*}{SNIPS}  \\
Model                           &                       &                         \\ 
\hline
Aligned LSTM with attention & 98.30                 & 98.10                   \\ 
\hline
Regular Transformer decoder     & 97.42                 & 97.37                   \\ 
\hline
Aligned Transformer decoder     & 98.40                 & 98.21                   \\
\hline
\end{tabular}
			\caption{\label{table_notaligned} The efficacy of TD alignment. Numbers show the SF f1 score of each model. Experiments were conducted on $BERT_{base}$. }
		\end{table}
	\subsubsection{Language Model Incorporation Strategy}
		Table \ref{table_arch_accuracy} compares two strategies: language model as an encoder and language model as word embeddings. In the former, we only use CTRAN's decoder after the language model. In the latter, we use both encoder and decoder of the CTRAN. Also, we experiment with two different language models for each strategy. Using ELMo as the encoder with CTRAN's decoder already surpasses previous SF state-of-the-art on ATIS. We observe an increase in SF f1 and ID accuracy when ELMo is used as word embeddings instead of an encoder. 
		Using BERT\textsubscript{base} as word embeddings with CTRAN, had a better performance on SNIPS when compared to BERT\textsubscript{base} as encoder. In contrast, Language model as word embedding did not do well on ATIS as it did not cause any improvement in results. It may be because ATIS is a smaller dataset; thus, having additional network layers can cause the final model to overfit.
		For BERT\textsubscript{large} Language model as word embeddings did well on both datasets. Furthermore, our architecture with BERT\textsubscript{large} achieved maximum performance. Although our results indicate that the complete architecture shows superior performance, it introduces extra computational costs. See Appendix \ref{sec:appendix_comp_burden} for details.
		Our experiments show that although BERT performs better related to ELMo in all cases, ELMo can be used for domain-specified datasets with nearly the same accuracy while having lower computational complexity and training time. This case may be due to goal-driven datasets not having a diverse vocabulary, thus making the presence of a pre-trained language model less significant. Also BERT\textsubscript{base} and BERT\textsubscript{large} do not have an advantage over each other in our implementation for intent-detection task. Their difference is shown in SF, where BERT\textsubscript{large} shows superiority over BERT\textsubscript{base} for all datasets.
		\begin{table*}
			\def\arraystretch{1.3}
\centering
\resizebox{\textwidth}{!}{\begin{tabular}{|l|l|c|c|c|c|} 
\cline{2-6}
\multicolumn{1}{c|}{}                                                                        & Dataset                        & \multicolumn{2}{c|}{ATIS} & \multicolumn{2}{c|}{SNIPS}  \\ 
\hline
Strategy                                                                                     & Model                          & ID acc & SF f1            & ID acc & SF f1              \\ 
\hline
\multirow{3}{*}{\begin{tabular}[c]{@{}l@{}}Language model\\as\\encoder\end{tabular}}         & ELMo + CTRAN's decoder         & 97.54  & 98.17            & 97.25  & 96.01              \\ 
\cline{2-6}
                                                                                             & BERT \textsubscript{base} 
 + CTRAN's decoder  & 97.99  & 98.44            & 98.86  & 98.00              \\ 
\cline{2-6}
                                                                                             & BERT \textsubscript{large} 
 + CTRAN's decoder & 97.99  & 98.43            & 98.86  & 98.18              \\ 
\hline
\multirow{3}{*}{\begin{tabular}[c]{@{}l@{}}Language model\\as\\word embeddings\end{tabular}} & CTRAN + ELMo                   & 97.88  & 98.25            & 97.73  & 96.68              \\ 
\cline{2-6}
                                                                                             & CTRAN 

+ BERT \textsubscript{base}            & 97.95  & 98.40            & 99.42  & 98.21              \\ 
\cline{2-6}
                                                                                             & CTRAN + BERT \textsubscript{large}             & 98.07  & 98.46            & 99.13  & 98.30              \\
\hline
\end{tabular}}
			\caption{\label{table_arch_accuracy} Two strategies used for finding the most suitable model. We used F-score for SF and accuracy for the ID task.}
		\end{table*}
	\section{Conclusion and Future Work}
		In this paper, we proposed CTRAN a novel CNN-Transformer-based architecture for joint intent-detection and slot-filling. The proposed model uses the encoder-decoder architecture. We use BERT as word embeddings and apply CNN with the window feature sequence structure to fuse local semantic information. Next, stacked Transformer encoders are used to provide the final encoder output. For the intent-detection task, we used self-attention followed by a fully connected layer. Additionally, We used a stack of aligned Transformer decoders for the slot-filling decoder. We compared our network with the well-known models, and the results show that our proposed CNN-Transformer model achieves state-of-the-art on slot-filling task.
		We also show that using language models as word embeddings is a better strategy than incorporating them into the structure. Using additional encoding after the language model introduces new parameters to the network, which comes with a small computational cost at training time and negligible inference delay.
		In the future, possible solutions to directly integrate predicted intent to slots and vice-versa can be explored in our architecture. Furthermore, we used pre-trained embeddings in a CNN-Transformer architecture. Future research can train a CNN-Transformer language model from scratch and remove the additional encoder altogether. Also, a more thorough language model comparison might be necessary.
		
	\bibliography{custom}
	\appendix
	\section{Computational Burden}
	\label{sec:appendix_comp_burden}
		The computational burden of additional encoding after the language model is a valid concern since more parameters are introduced to the network. Noting that we used a single RTX 3080 10GB for our computations, Table \ref{table_train_time} shows training time before and after using additional encoding. Our measurements show that between 6-14\% extra time is needed to converge the model. For this calculation, we ran each model 10 times for 10 epochs and reported the median value. 
		Table \ref{table_inference_time} shows the inference time of the model. For this experiment, we ran the model 10 times for 200 instances and reported the mean values. Albeit a small gain in training time, the increase in inference time is negligible. Comparing before and after using additional encoding, we observe between 1-1.5\% extra delay in inference. We anticipated this increase in training and inference time, since it is reasonable for a model with more parameters to have an extra delay.
		\begin{table}
			\def\arraystretch{1.4}
\centering
\begin{tabular}{|l|c|c|c|c|} 
\hline
\multicolumn{1}{|r|}{Dataset} & \multicolumn{2}{c|}{ATIS}           & \multicolumn{2}{c|}{SNIPS}           \\ 
\cline{2-5}
LM                            & LM as En. & Additional En. & LM as En. & Additional En.  \\ 
\hline
ELMo                          & 42.51         & 47.55 (11\%)        & 92.37         & 103.51 (11\%)        \\ 
\hline
BERT \textsubscript{base}     & 42.40         & 46.71 (10\%)        & 110.39        & 119.50 (8\%)         \\ 
\hline
BERT \textsubscript{large}    & 73.57         & 79.44 (7\%)         & 182.42        & 194.24 (6\%)         \\
\hline
\end{tabular}

			\caption{\label{table_train_time}Additional training time caused by the proposed CNN-Transformer encoder. Values show how many seconds it takes for the entire model to be trained for one epoch over the noted dataset. The value in parentheses indicates the percentage increase. LM is short for Language Model, and En. indicates encoding.  }
		\end{table}  
		\begin{table}
			\def\arraystretch{1.4}
\centering
\begin{tabular}{|l|c|c|c|c|} 
\hline
\multicolumn{1}{|r|}{Dataset} & \multicolumn{2}{c|}{ATIS}           & \multicolumn{2}{c|}{SNIPS}           \\ 
\cline{2-5}
LM                            & LM as En. & Additional En. & LM as En. & Additional En.  \\ 
\hline
ELMo                          & 208           & 211 (1.5\%)          & 192           & 195 (1.5\%)           \\ 
\hline
BERT \textsubscript{base}     & 178           & 181 (1.5\%)        & 177           & 179 (1\%)           \\ 
\hline
BERT \textsubscript{large}    & 186           & 189 (1.5\%)        & 184           & 187 (1.5\%)         \\
\hline
\end{tabular}
			\caption{\label{table_inference_time}Comparing inference time before and after additional encoding is applied. Values show how many milliseconds it takes for a single example to be inferred. The value in parentheses indicates the percentage increase. LM is short for Language Model, and En. indicates encoding. }
		\end{table}
\end{document}